\documentclass[]{interact}

\usepackage{epstopdf}
\usepackage[caption=false]{subfig}
\usepackage{tabularx}
\usepackage{hyperref}
\usepackage{algorithm} 
\usepackage{subfig}
\usepackage{algpseudocode} 
\usepackage{graphicx}               
\usepackage{amsmath,amssymb}
\usepackage[font=small,skip=4pt]{caption}
\usepackage{caption}
\usepackage{parskip}
\usepackage[compact]{titlesec}
\usepackage{array}
\newcolumntype{P}[1]{>{\centering\arraybackslash}p{#1}}
\usepackage[justification=centering]{caption}
\usepackage[utf8]{inputenc}
\usepackage[para,online,flushleft]{threeparttable}
\usepackage{xcolor}
\usepackage{natbib}
\bibpunct[, ]{(}{)}{;}{a}{}{,}
\renewcommand\bibfont{\fontsize{10}{12}\selectfont}

\theoremstyle{plain}

\theoremstyle{definition}

\theoremstyle{remark}

\begin{document}
\title{A semi-supervised Teacher-Student framework for surgical tool detection and localization}

\author{
\name{Mansoor Ali\textsuperscript{a} \thanks{CONTACT Mansoor Ali. Email: a01753093@tec.mx}, Gilberto Ochoa-Ruiz\textsuperscript{a}, and Sharib Ali\textsuperscript{b}}
\affil{\textsuperscript{a}Tecnologico de Monterrey, Escuela de Ingeniería y Ciencias, 38115, México; \textsuperscript {b}School of Computing, University of Leeds, Leeds, UK}
}

\maketitle

\begin{abstract}
Surgical tool detection in minimally invasive surgery is an essential part of computer-assisted interventions. Current approaches are mostly based on supervised methods which require large fully labeled data to train supervised models and suffer from pseudo label bias because of class imbalance issues. However large image datasets with bounding box annotations are often scarcely available. Semi-supervised learning (SSL) has recently emerged as a means  for training large models using only a modest amount of annotated data; apart from reducing the annotation cost. SSL has also shown promise to produce models that are more robust and generalizable. Therefore, in this paper we introduce a semi-supervised learning (SSL) framework in surgical tool detection paradigm which aims to mitigate the scarcity of training data and the data imbalance through a knowledge distillation approach. In the proposed work, we  train a model with labeled data which initialises the Teacher-Student joint learning, where the Student is trained on Teacher-generated pseudo labels from unlabeled data. We propose a multi-class distance with a margin based classification loss function in the region-of-interest head of the detector to effectively segregate foreground classes from background region. Our results on m2cai16-tool-locations dataset indicate the superiority of our approach on different supervised data settings (1\%, 2\%, 5\%, 10\% of annotated data) where our model achieves  overall improvements of 8\%, 12\%  and 27\% in mAP (on 1\% labeled data) over the state-of-the-art SSL methods and a fully supervised baseline, respectively.  The code is available at \textcolor{blue}{\url{https://github.com/Mansoor-at/Semi-supervised-surgical-tool-detection}}. 
\end{abstract}

\section{Introduction}
Recent works in deep learning based on visual recognition methods have delivered enormous advantages towards computer-assisted interventions (CAI)~\citep{ward2021computer}. CAI tools have been primarily focused on specific information gathering, such as the presence or location of lesions. Nonetheless, recent developments in the image recognition tasks with improved accuracy have led to expansion of its scope to several other areas including intra-operative decision support systems~\citep{bouget2017}. These applications provide contextual information to the surgeon during the surgery, as a post-operative feedback~\citep{bhatia2007,sarikaya2017} for surgical training and video content analysis~\citep{wang2019graph}. 

More recently, CAI systems capable of performing effectively the sub-tasks such as surgical phase recognition, identification of the tool presence, as well as their recognition, localization and instance-based segmentation are getting increased attention~\citep{bouget2017}. The development of these task-based automated approaches can ensure improved surgical care, patient safety and alleviate surgeon fatigue. 


Deep learning based surgical tool detection task has attracted a lot of attention in recent years. However, most of the state-of-the-art (SOTA) methods have employed fully supervised approaches~\citep{jin2018, zhang2020} and only a few weakly supervised methods, mostly implementing classification models for determining tool presence~\citep{vardazaryan2018} have been proposed. Nonetheless, training complex deep learning (DL) models under the supervised setting requires difficult-to-acquire and precisely annotated datasets, which is a time consuming task and susceptible to intra and inter-observer bias in annotations. As a result, only a few labeled surgical tool datasets are publicly available~\citep{sarikaya2017,jin2018} and this lack of annotated datasets has essentially hindered the development of robust and generalizable deep architectures for the surgical instrument detection.




\begin{figure}%
    \captionsetup{justification=justified}
    \subfloat[\centering]{{\includegraphics[scale=0.4]{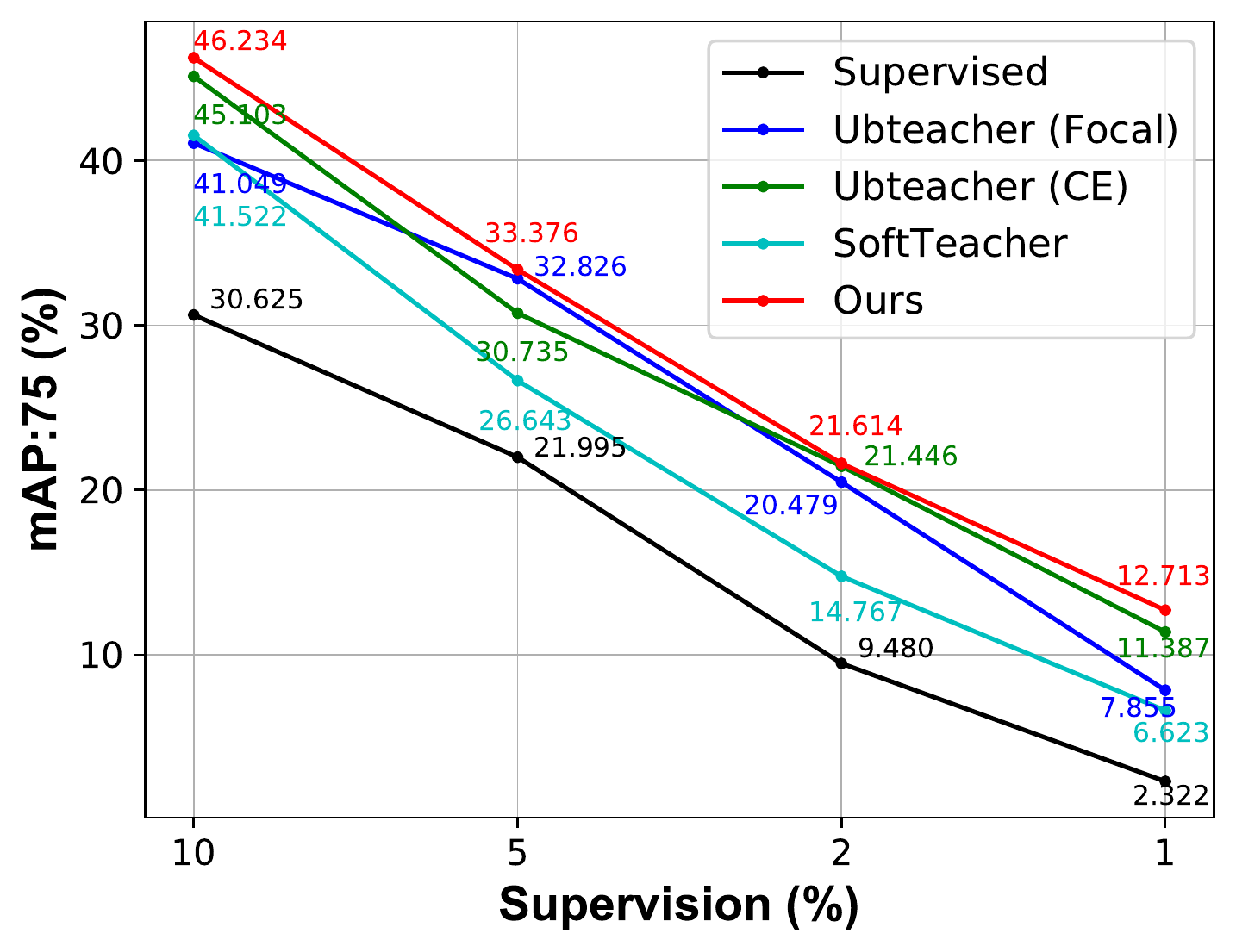} }}%
    \subfloat[\centering]{{\includegraphics[scale=0.52]{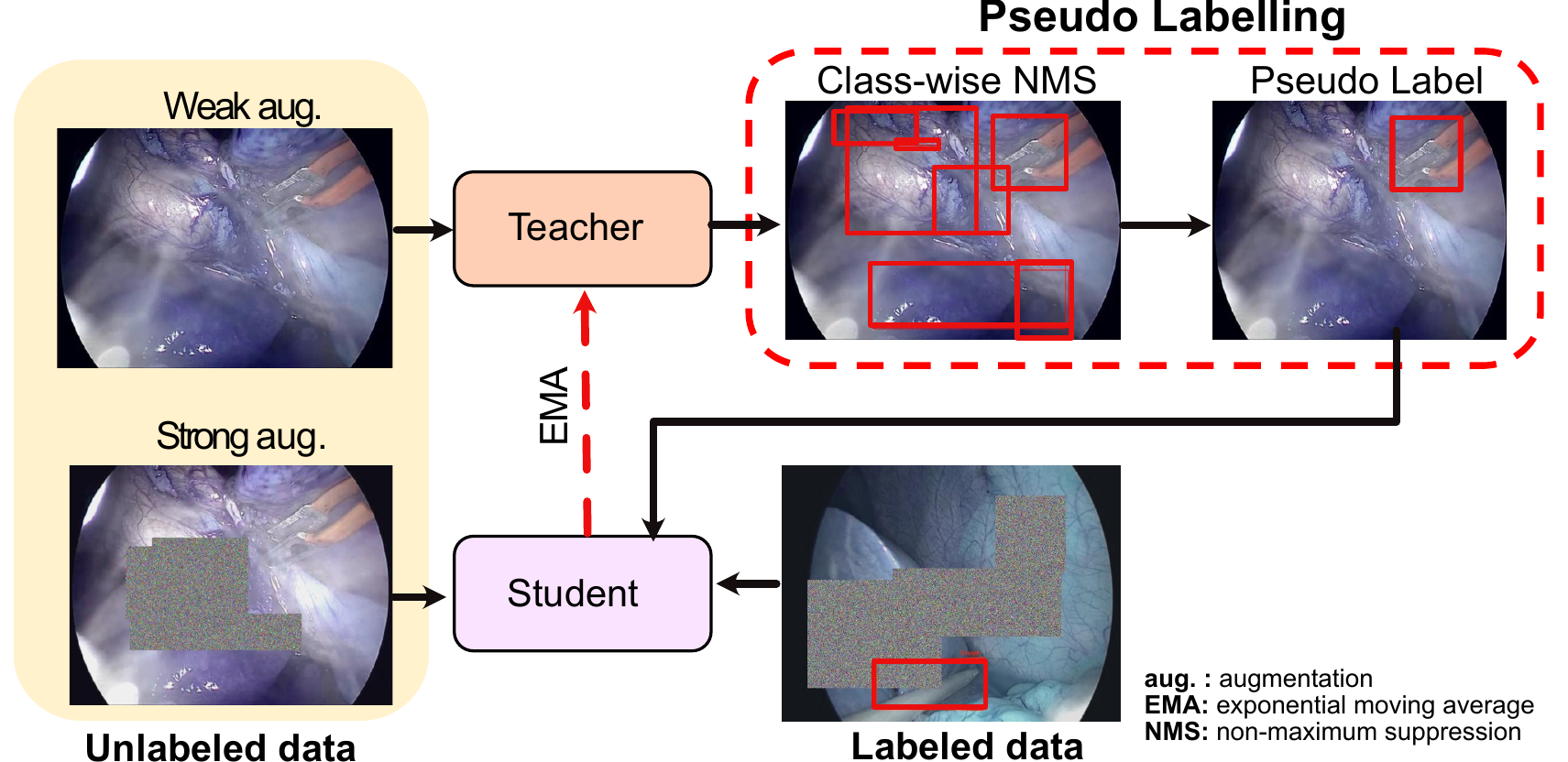}}}%
    \caption{\textbf{Comparison at different percentage of supervision and augmentation strategies in a Teacher-Student paradigm.} (a) The proposed approach efficiently leverages unlabeled data and produces substantial improvement over supervised baseline and on the SOTA Unbiased Teacher (Ubteacher)~\cite{unbiased} framework. (b) Data augmentation workflow of the proposed Teacher-Student mutual learning approach. Unlabeled data given with weak augmentation to Teacher and with strong augmentation to Student. Student gets pseudo labels through NMS and thresholding from the Teacher. }%
    \label{Overall result and augmentation}%
\end{figure}
Alternatively, the annotation cost could be greatly mitigated by exploiting unlabeled data through efficient semi-supervised learning (SSL) frameworks. The core idea of SSL is to be able to extract information from the unlabeled data that is essential for label prediction. One solution is to train a network to solve a pre-defined pretext task (Teacher model generating pseudo labels) and then using the learned knowledge in the downstream task (Student network). Recently, SSL has shown promising outcomes in improving model performance and is receiving growing attention of the computer vision research community \citep{van2020,sohn2020fixmatch}. Despite these progresses, most of these advances are in the domain of image classification rather than object detection as the bounding box annotations require more time and effort to generate. Traditionally, SSL can be approached with adapting SOTA image classification methods such as \citep{sohn2020fixmatch} to object detection. However, existence of some unique characteristics such as foreground-background and foreground class imbalance makes object detection interact poorly with those methods. The class imbalance problem may greatly impede the use of pseudo-labeling based training pipelines since Teacher generated pseudo labels will be overly biased towards dominant classes and ignoring minor and less dominant classes. As a result, these models in their vanila arrangement will exacerbate class-imbalance problem and cause severe overfitting. 



To overcome these issues, we propose a jointly trained Teacher-Student model on m2cai16-tool-locations dataset~\citep{jin2018} which is initialised by a supervised detector. We argue that slowly updating the Teacher by exponential moving average (EMA) via the Student can alleviate pseudo-labeling bias problem and improve pseudo label quality, hence overall performance improvement. Additionally, we propose a multi-class distance and margin-based classification loss in the ROI head of the detector network to boost the classification performance. This is achieved by maximising the distance between foreground classes and the background. To the best of our knowledge, our approach is the first effort towards leveraging Teacher-Student joint training paradigm for addressing data scarcity problem in surgical tool detection applications. We employ strong and weak augmentation pipelines to improve model robustness (Fig.~\ref{Overall result and augmentation}(b)). Our proposed pipeline outperforms supervised baseline and other SOTA semi-supervised methods in terms of classification and localisation performance (Fig.~\ref{Overall result and augmentation}(a)).    


In the rest of the paper, we discuss related work (section \ref{Literature Review}), materials and method (section \ref{Materials and Method}), quantitative and qualitative results (section \ref{results}), ablation studies (section \ref{ablation study}) and lastly discussion and conclusion (section \ref{conclusion}).




\section{Related Work}\label{Literature Review}
Some of the early works on surgical tool detection used radio frequency identification tags \citep{kranzfelder2013}, Viola-Jones detection algorithm~\citep{lalys2011} and segmentation, contour delineation and three-dimensional modeling \citep{speidel2009}. With the advent of deep learning-based approaches using convolutional neural networks, computer vision methods have evolved with remarkable growth and demonstrated promising outcomes~\citep{imagenet}. In the surgical domain, several works have leveraged deep learning approaches to obtain SOTA performance on surgical instrument detection~\citep{jin2018,sahu2016tool,twinanda2016,endonet}.

Most of the studies conducted on surgical tool detection have proposed supervised pipelines or only have targeted frame-level tool presence detection. For example, AGNet \citep{agnet} used global and local prediction networks to obtain visual cues for tool presence detection and showed a significant improvement over m2cai16-tool challenge \citep{raju2016m2cai} winners. Jin \emph{et al.}~\citep{jin2018} proposed region-based convolutional neural network to perform surgical skill assessment adapted to tool presence detection, spatial localization and tracking. The authors also extended the m2cai16-tool dataset \citep{dataset} to include tool bounding boxes (subsequently named as m2cai16-tool-locations) which we have used in this work. Sarikaya \emph{et al.} used image and temporal motion cues to train multi-modal CNN models \citep{sarikaya2017} for tool detection and localization in robotic-assisted surgical training task videos. Tool detection and pose estimation was also studied in \citep{reiter2012} but it was limited to robotic arms that return kinematic data. Shi \emph{et al.} proposed a lightweight attention-guided framework \citep{shi2020} for tool detection and conducted an ablation study on three different datasets (two public datasets, EndoVis
Challenge \citep{kurmann2017} and ATLAS Dione \citep{sarikaya2017} and one self-prepared cholec80-locations). 
However, their model performed well on all tools except grasper and irrigator classes. In another study \citep{zhang2020}, irrigator can be observed as worst performing instrument with average precision of 41.6\%, followed by grasper with 54.1\% in a supervised setting at IOU threshold of 50\%. 
A ghost feature maps-based pipeline was used to reduce the computational burden for tool detection in \citep{yang2021efficient}. A CNN-based hidden Markov model was proposed by Twinanda in \cite{twinanda2016} for surgical tool detection from laparoscopic videos. A combination of CNN to extract spatial features and long short-term memory (LSTM) for temporal cues was proposed to perform surgical tool detection from laparoscopic videos~\citep{mishra2017}. 

Although the results of some of these approaches have been mostly encouraging, they have reported only one mAP results \citep{sarikaya2017, shi2020:IEEE} which is not quite sufficient to gauge the classification and localisation performance. Furthermore, previous approaches require completely labeled datasets to train the model. Such datasets are either scarcely available or the process of annotating them can lead to other issues such as introducing unintended biases in the trained model.

In this work, we aim to demonstrate the advantages of an SSL approach and propose a novel semi-supervised Teacher-Student framework to alleviate the limited data problem and annotation cost requirement for training on larger datasets. Our literature search revealed that there are only two studies conducted on semi-supervised learning in the medical domain where one is based on cataract surgery dataset \citep{jiang2021semi} while another study \citep{yoon2020semi} used a tracker to detect instruments from unlabeled private surgery videos. To the best of our knowledge, this is the first approach that investigates the effectiveness of unlabeled data through a Teacher-Student learning pipeline for tool detection on a minimally invasive surgery dataset. We report results from our model in terms of mAP on various IOU thresholds to demonstrate the effectiveness of our approach in detecting and localising surgical tools. 


\section{Materials and Method}\label{Materials and Method}
\subsection{Dataset} \label{dataset}
In this work, we use an extended version of the m2cai16-tool dataset which was originally released for M2CAI 2016 Tool Presence Detection Challenge \citep{endonet}. This dataset consists of 15 videos each with duration from 20 to 75 minutes of cholecystectomy procedures performed at the University Hospital of Strasbourg in France. After down sampling at 1 fps, it leaves 23,000 frames annotated with tool presence classification.  

\begin{figure}[t!]
\captionsetup{justification=justified}
\includegraphics[width= \textwidth]{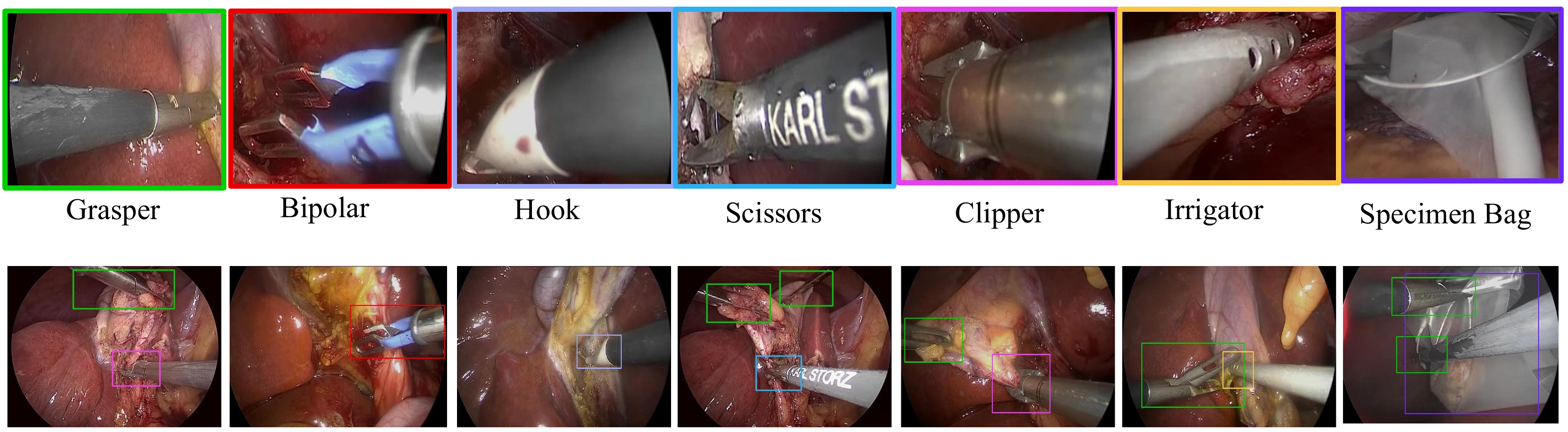}
\caption{(Top) Samples from \textit{m2cai16-tool-locations} dataset with representative classes of seven tools. (Bottom) Example frames with bounding box annotations where color of the box refers to tool class} \label{tools and labels}
\end{figure}

Later, m2cai16-tool-locations dataset was build with spatial bounding box annotations~\citep{jin2018}. This dataset consists of a total of 2812 frames that were annotated under supervision and spot-checking from clinical experts. We have used 80\%, 10\%, and 10\% for training, validation and test splits, respectively. The annotations breakdown per class is given in \textbf{supplementary material (Table 1)} and the tool instances with example box annotations are presented in Fig.~\ref{tools and labels}. 

We use average precision (AP) computed per class and mean average precision (mAP) for all seven classes which are the standard object detection evaluation metric. These metrics are evaluated at different IoU thresholds, usually denoted as \(mAP_{IoU- threshold}\). We report results for 50, 75, 50:95 (average of AP values for IoU thresholds from 50 to 95 with interval of 5), medium and large IoU thresholds.



%
%

\subsection{Data Augmentation}
We have used two data augmentation strategies in this work, which we refer as weak and strong augmentations (Fig.~\ref{Overall result and augmentation}(b)). For the weak augmentation, we apply random horizontal flips whilst for strong augmentation, we randomly perform several photometric augmentations like grayscale, color jittering, Gaussian blur, patch masking and cutout patches~\cite{devries2017}. For the complete description of data augmentation with parameter values, please refer to \cite{unbiased}. 

\section{Method} \label{method}
\begin{figure}[h!]
\captionsetup{justification=justified}
\includegraphics[width= \textwidth, keepaspectratio]{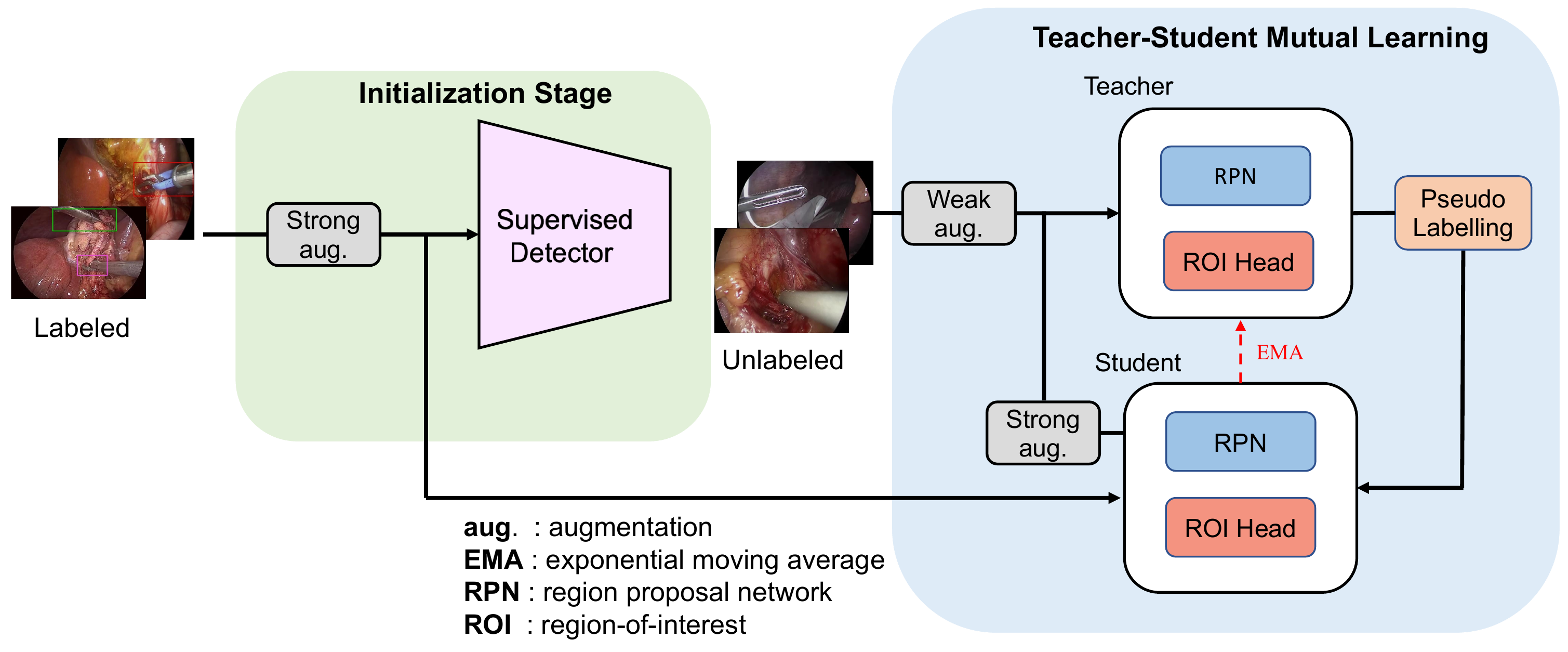}
\caption{Overview of the proposed Surgical tool detection model. It consists of two modules: 1) An initialisation module, where a supervised model makes use of strongly augmented labeled data, 2) A Teacher-Student mutual learning module, where the Student is trained with strongly augmented unlabaled data with Teacher-generated pseudo labels. The Student transfers learned weights to the Teacher gradually through Exponential Moving Average (EMA).} \label{General Block DIagram}
\end{figure}

In this work we address multi-instance surgical tool detection problem in a semi-supervised setting. Let the training set in various arrangements of labeled data sets be denoted as \(D_s = \{x_i^s, y_i^s\}_{i=1}^{N_s}\) and unlabeled data sets be \(D_u = \{x_i^u\}_{i=1}^{N_u}\), where \(N_s\) and \(N_u\) represent number of supervised and unsupervised training samples while \(y^s\) represent bounding box annotation of each labeled image \(x^s\). Here, \(y^s\) consists of bounding boxes for all object instances, height and width of image and instance category names. It is important to mention that since all the training data samples contain labels, during training we removed the labels of the portion we categorise as unlabeled. The overall training pipeline is divided into two stages as shown in Fig.~\ref{General Block DIagram}. The first stage is the initialization stage (section \ref{initial}), while the second is the Teacher-Student joint learning mechanism (section~\ref{joint learning}). In the second stage, the Teacher generates pseudo-labels and Student network is trained on both pseudo labeled data and supervised data. Each stage is detailed separately below along with Student learning and Teacher update scheme and  margin . 
%
\subsection{Initialization stage} \label{initial}
The initialization stage acts as a trigger point for Teacher-Student joint learning. It sets the stage for the Teacher model to be able to generate qualitative pseudo-labels for better Student learning. In this stage, we exploit the available labeled data  \(D_s = \{x_i^s, y_i^s\}_{i=1}^{N_s}\) to train the  Faster-RCNN detector model (\(\theta\)) with supervised loss \( \mathcal{L}_{sup} \). The standard Faster-RCNN model makes use of four losses: RPN classification loss \( \mathcal{L}_{cls}^{rpn} \), RPN regression loss \( \mathcal{L}_{reg}^{rpn} \), ROI classification loss \( \mathcal{L}_{cls}^{roi} \) and ROI regression loss \( \mathcal{L}_{reg}^{roi} \) (Eq. (1)). 
\begin{equation}
   \mathcal{L}_{sup} = \sum_i^{N_s}   \mathcal{L}_{cls}^{rpn} (x_i^s, y_i^s) + \mathcal{L}_{reg}^{rpn} (x_i^s, y_i^s) + \mathcal{L}_{cls}^{roi} (x_i^s, y_i^s) + \mathcal{L}_{reg}^{roi} (x_i^s, y_i^s)
\end{equation}

The weights and architecture of the model trained during this initialization phase are then copied to be used for both the Student and Teacher models \begin{math} (\theta_\mathcal{T} \leftarrow \theta, \theta_\mathcal{S} \leftarrow \theta) \end{math}. The trained detector from this stage provides a good initialization for next stage, where we further exploit unsupervised data to improve object detection. 

\subsection{Teacher-Student joint learning stage} \label{joint learning}

The proposed knowledge distillation framework leverages Student and Teacher joint training to address lack of data problem. During training, Teacher generates pseudo labels on unlabeled data and Student is trained on those labels. Thus, a continuously learning Student passes on the learned knowledge to the Teacher. We posit that this evolving mutual learning would result in better detection performance by generating stable and reliable pseudo labels. Weak and strong augmentation pipelines ensure reliable pseudo label generation by Teacher and diversity in Student models respectively. 


\subsection{Student learning and Teacher update scheme}
We tackle the pseudo-label noise problem which may cause severe performance degradation \citep{sohn2020fixmatch} by confidence thresholding (\(\tau\)). Although this step could have sufficed in the case of image classification, for object detection tasks, additional steps must be enforced as duplicated bounding box predictions and class imbalanced prediction problems are typically encountered in these settings.  We address the duplicated box predictions problem by applying class-wise non-maximum suppression (NMS) before a confidence thresholding step. As simple confidence thresholding only removes samples with low confidence on predicted object categories and does not take into account the quality of bounding box locations, we do not use unsupervised loss on bounding box regression which is thus represented as below with $\theta_{S}$ as weight updates between both supervised $\mathcal{L}_{sup}$ and unsupervised $\mathcal{L}_{unsup}$ losses: 

\begin{align}
   \mathcal{L}_{unsup} = \sum_i^{N_u}   \mathcal{L}_{cls}^{rpn} (x_i^u, \Tilde{y}_i^u) + \mathcal{L}_{cls}^{roi} (x_i^u, \Tilde{y}_i^u)\\
      \theta_{\mathcal{S}} \leftarrow \theta_{\mathcal{S}} + \gamma \frac{\partial(\mathcal{L}_{sup} + \lambda_u \mathcal{L}_{unsup})}{\partial\theta_\mathcal{S}},
\end{align}
\noindent{where} \(\gamma\) is the learning rate and \(\lambda_u\) is unsupervised loss weight. The overall unsupervised loss in Eq. (2) consists of the sum of RPN and ROI head classification losses. Eq. (3) depicts the Student weight update scheme which includes both supervised and unsupervised losses with a loss weight parameter $\lambda_u$. 


Finally, we perform Teacher model refinement by using EMA following \emph{Mean Teacher} to slowly update Teacher network which in turn will generate stable and reliable pseudo labels. The update can be represented as:
\begin{equation}
   \theta_{\mathcal{T}} \leftarrow \alpha\theta_{\mathcal{T}} + (1- \alpha)\theta_{\mathcal{S}}, 
\end{equation}
\noindent{where} \(\alpha\) is the EMA rate, and \(\theta_\mathcal{T}\), \(\theta_\mathcal{S}\) are the network weights for Teacher and Student.

\subsection{Logistic loss with added margin and distance penalization}
In the surgical domain, foreground class imbalance exists in every dataset due to the fact that tool usage frequency varies from one tool to another \citep{mishra2017}. In this work, to address the class imbalance problem, we target the foreground and background class imbalance problem by introducing a multi-class loss function based on a margin, which tries to maximise foreground-background distance. Unlike the focal or cross entropy losses, our proposed loss tries to predict relative distance between inputs. Specifically, we divide classification logits between foreground and background instances and then compute \textit{sigmoid} probability, respectively. We then sum the \textit{softmax} of the probabilities over all the batch for the foreground \(\rho\) and background \(\beta\) instances. These probabilities are then used to maximise foreground-background distance in the final loss computation which is in the form of a logistic loss function for classification defined as:



%
\begin{equation}{\label{eq:5}}
   \mathcal{L}_{cls}^{roi} = \sum_n w_l \log(1+ \frac{e^{ \textstyle s\cdot(\beta-\rho+\sigma)}}{s}),
\end{equation}
\noindent{where} $n$ is the mini-batch size, \(w_l\) represents loss weight, s is the smoothness parameter and \(\sigma\) denotes margin. 

Apart from the multi-class loss, Teacher update with EMA will also help reduce pseudo label bias since new Teacher is regularised by previous Teacher model which prevents drastic movement of the decision boundary towards under-represented classes. 


\begin{algorithm}
\caption{Multi-class distance and margin based classification loss}\label{loss}

\begin{algorithmic}[1]
\Procedure{loss}{$logits, targets$}
\State $ classes \gets  class\_indices$
\State $fg\_logits \gets logits(targets = classes)$
\State $ bg\_logits \gets  logits(targets != classes)$
\State $fg\_prob \gets sigmoid(fg\_logits)$
\State $bg\_prob \gets sigmoid(bg\_logits)$
\State $\textcolor{blue}{\rho} \gets \sum softmax(fg\_prob)$
\State $\textcolor{blue}{\beta} \gets \sum softmax(bg\_prob)$
\State $ loss \gets  Eq:5 $

\EndProcedure
\end{algorithmic}
\end{algorithm}

\section{Experiments and results}\label{results}
\subsection{Implementation Details}
The implementation of our proposed framework is based on Faster-RCNN detector model with ResNet50-FPN bachbone, whose network weights are initialized by ImageNet pretrained model. We use a confidence threshold (\(\tau\)) of 0.7, regularization co-efficient for unsupervised loss (\(\mathcal{\lambda}_u\)) of 0.2 and EMA rate (\(\alpha\)) of 0.9996. We use \textit{WarmupMultiStepLR} as a learning rate (\(\alpha\)) scheduler in initialization stage while a constant learning rate of 0.01 for the Teacher-Student mutual learning stage. In the initialization stage, we use strong augmentation, while during the Teacher-Student mutual learning, we use weak augmentation for the Teacher and strong augmentation for Student. We report results in terms of mAP on different IOU thresholds. We use a batch size of 8 (4 labeled images and 4 unlabeled images) throughout the experiments. We performed network training through detectron2 \citep{wu2019detectron2} object detection framework using 4 GPUs on NVIDIA Tesla P100-SXM2-16GB system. We use fixed seed values for generating the data splits to make the results more reproducible.
\subsection{Results}
\subsubsection{Quantitative Results}
We evaluate our model with different labeled and unlabeled data protocols and present the results on a 10\% held-out set in Table \ref{Experimental results}. The table also includes results on the supervised baseline, UnbiasedTeacher \citep{unbiased} with both CrossEntroppy and focal losses. and SoftTeacher \citep{softteacher}. Table \ref{perclassAP} shows per class \(mAP_{50:95}\) results on 1\% labeled data setting. Furthermore, we also conduct a paired t-test between \(AP_{50}\) obtained by our proposed model and  \(AP_{50}\) obtained by other SOTA methods. The resulting box-plot on 1\%, 2\%, 5\% and 10\% labeled data setting are given Fig.\ref{fig:box-plot} and \textit{p}-values are shown in Table \ref{Experimental results}.   
%
\begin{table}[t!]
\tbl{Experimental results with ResNet50-FPN as backbone .}
{\begin{tabular}{lcccccc} \toprule
 & \multicolumn{5}{c}{ \textbf{1\% Labeled data}}& \textit{p}-{values} \\ \cmidrule{2-6}
Method& $mAP_{50}$ & $AP_{50:95}$& $mAP_{75}$& $mAP_{m}$& $mAP_{l}$ \\ \midrule
Supervised & 23.578 & 7.673 & 2.322 & 6.189 & 9.050 & $5.996e-17$  \\

Unbiased Teacher\textsuperscript{$*$} \citep{unbiased} & 34.374&14.145& 7.855 &10.687 & 15.880 & $5.626e-02$ \\
Unbiased Teacher\textsuperscript{$**$} \citep{unbiased} & 42.382    &  18.008  & 11.387 & 13.041 & 20.135 & $6.229e-03$   \\
SoftTeacher \citep{softteacher} & 38.421    &  13.556  & 6.623 & 16.756 & 13.045 & $5.526e-02$  \\
Ours & 50.632 & 20.094 & 12.713 & 15.219 & 21.774 & -- \\ \bottomrule
& \multicolumn{5}{c}{ \textbf{2\% Labeled data}} \\ \cmidrule{2-6}
Supervised & 47.140 & 18.609 & 9.480 & 24.033 & 18.586 & $2.558e-14$  \\

Unbiased Teacher\textsuperscript{$*$} \citep{unbiased} & 71.608 & 31.752 & 20.479 & 27.871 & 32.430  & $3.975e-04$ \\
Unbiased Teacher\textsuperscript{$**$} \citep{unbiased} & 72.416 & 31.490 & 21.446 & 26.767 & 32.666 & $2.010e-01$  \\
SoftTeacher \citep{softteacher} & 60.366    &  25.421  & 14.767 & 17.991 & 28.323 & $2.558e-8$  \\
Ours & 72.341 & 32.311 & 21.614 & 29.780 & 33.556 & -- \\ \bottomrule
& \multicolumn{5}{c}{ \textbf{ 5\% Labeled data}} \\ \cmidrule{2-6}
Supervised & 71.082 & 32.249 & 21.995 & 29.505 & 35.041 & $5.866e-05$  \\
Unbiased Teacher\textsuperscript{$*$} \citep{unbiased} & 84.721 & 42.269 & 32.826 & 35.697  & 44.204 & $1.298e-01$ \\
Unbiased Teacher\textsuperscript{$**$} \citep{unbiased} & 82.592 & 40.393 & 30.735 & 33.665 & 42.904 & $3.424e-04$ \\
SoftTeacher \citep{softteacher} & 83.211   &  38.857  & 26.643 & 30.567 & 40.718 & $4.566e-04$ \\
Ours & 84.427 & 42.392 & 33.376 & 31.156 & 44.610 & -- \\ \bottomrule
& \multicolumn{5}{c}{ \textbf{10\% Labeled data}} \\ \cmidrule{2-6}
Supervised & 80.193 & 38.640 & 30.625 & 29.845 & 40.958 & $7.108e-03$  \\
Unbiased Teacher\textsuperscript{$*$} \citep{unbiased} & 92.981 & 47.369 & 41.049 & 41.137 & 48.714 & $2.291e-01$ \\
Unbiased Teacher\textsuperscript{$**$} \citep{unbiased} & 90.353 & 45.972 & 45.103 & 39.247 & 47.787 & $1.00$ \\
SoftTeacher \citep{softteacher} & 89.362   &  42.717  & 41.522 & 38.312 & 43.849 & 1.00  \\
Ours & 90.250 & 46.886 & 46.234 & 42.635  & 48.644 & --  \\ \bottomrule
\end{tabular}}
\tabnote{\textsuperscript{*} with Focal Loss ; \textsuperscript{**} with Cross Entropy Loss.}
\label{Experimental results}
\end{table}
\begin{table}[t!]
\centering
\begin{threeparttable}[b]
\caption{The average precision (\(AP_{50:95}\)) per class on 1\% labeled data.}\label{perclassAP}

\footnotesize
\begin{tabular}{p{2cm}P{1.3cm}P{1.6cm}P{1.7cm}P{1.5cm}P{1.5cm}}
\midrule
Class & Supervised & Ubteacher\textsuperscript{*}  & Ubteacher\textsuperscript{**}&  SoftTeacher &    Ours  \\ \midrule
Grasper &  12.434      &  23.457 & 23.046     &  7.0  &  20.203  \\
\midrule
Bipolar &    13.450     & 19.472 & 33.384    &   22.2    &  29.499 \\
\midrule
Hook &   11.349          & 38.529 & 44.614     &   29.8     &    43.924 \\
\midrule
Scissors &  3.592           & 4.130 & 5.052    &   3.8    &   6.860  \\
\midrule
Clipper &  4.273           & 5.045    & 4.800    &    0.0    &   4.970 \\
\midrule
Irrigator &  4.022           &  3.393 & 8.468     &   27.9     &    10.331  \\
\midrule
SpecimenBag &  4.592         & 4.986  & 6.692        &    4.1     &   24.873     \\
\bottomrule
\end{tabular}
\begin{tablenotes}
   \item[*] Unbiased Teacher with focal loss;  \item[**] Unbiased Teacher with cross entropy loss;  
\end{tablenotes}
\end{threeparttable}
\end{table}

\begin{figure}[t!]
\captionsetup{justification=justified}
\centering
\includegraphics[width=\textwidth,height=3in, keepaspectratio]{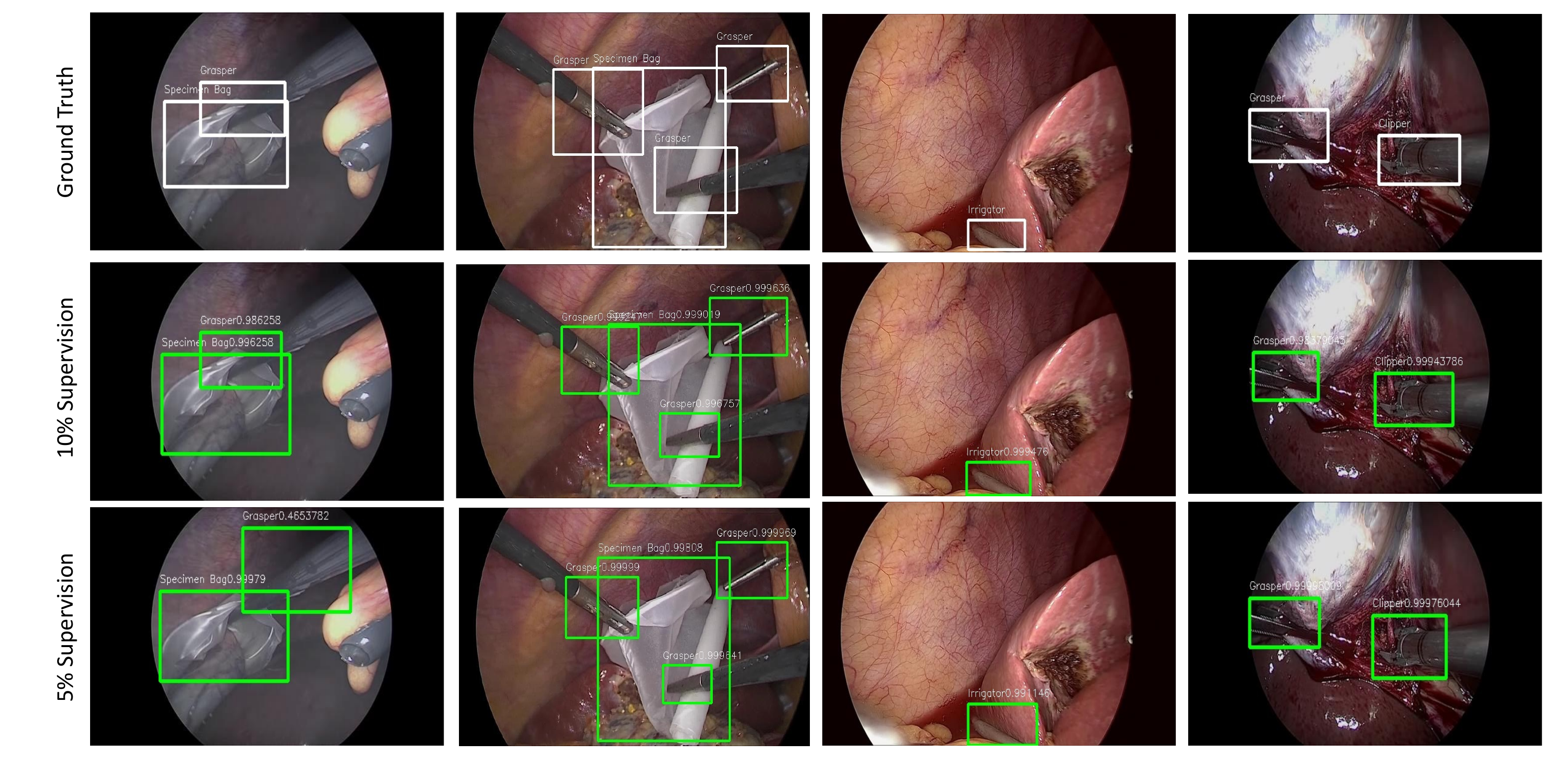}
\includegraphics[width=\textwidth,height=3in, keepaspectratio]{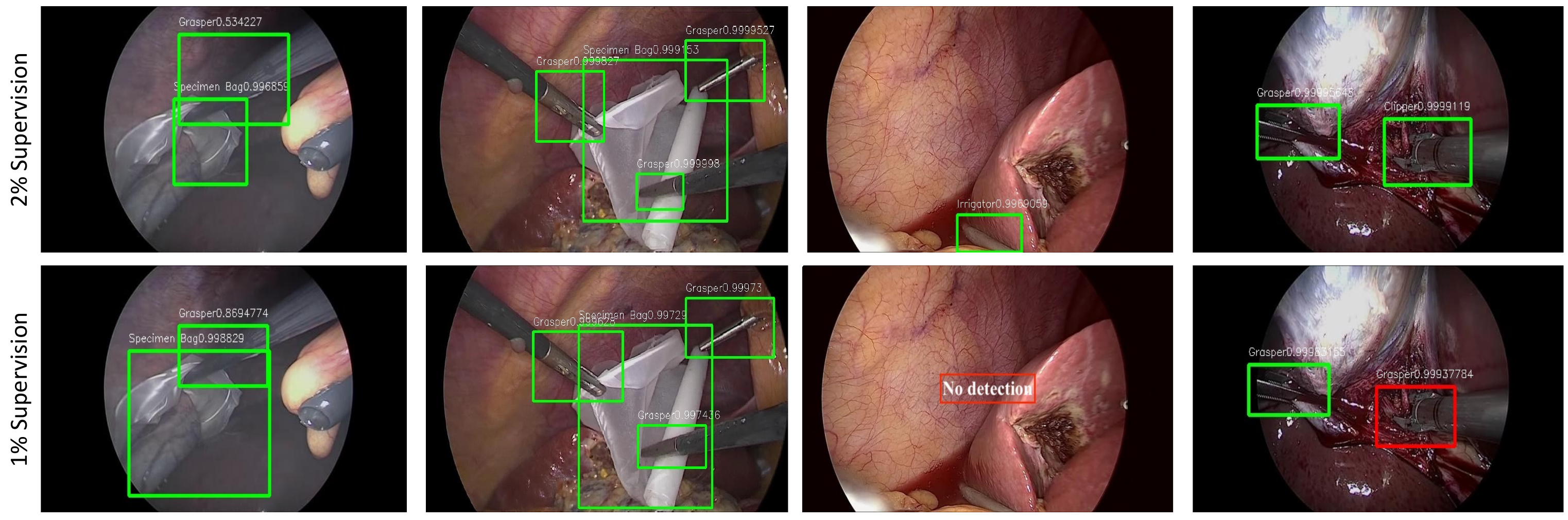}
\caption{Qualitative Results: First row shows images with ground truth. Second, third, fourth and fifth row presents results on 10\%, 5\%, 2\% and 1\% setting respectively, Green and red boxes indicate correct and wrong predictions} \label{qualitative}
\end{figure}
\subsubsection{Qualitative Results}
In this section, we report the qualitative performance of our model as shown in Fig.~\ref{qualitative}. The example surgical scenes are carefully chosen to contain several instances in one frame (column two from left), only partially visible instrument (column three from left), irregular orientation (column four from left). Results on all data settings have been presented to see how well model performs in terms of detection and localisation.

\begin{figure}[t!]
\centering
\includegraphics[width= \textwidth]{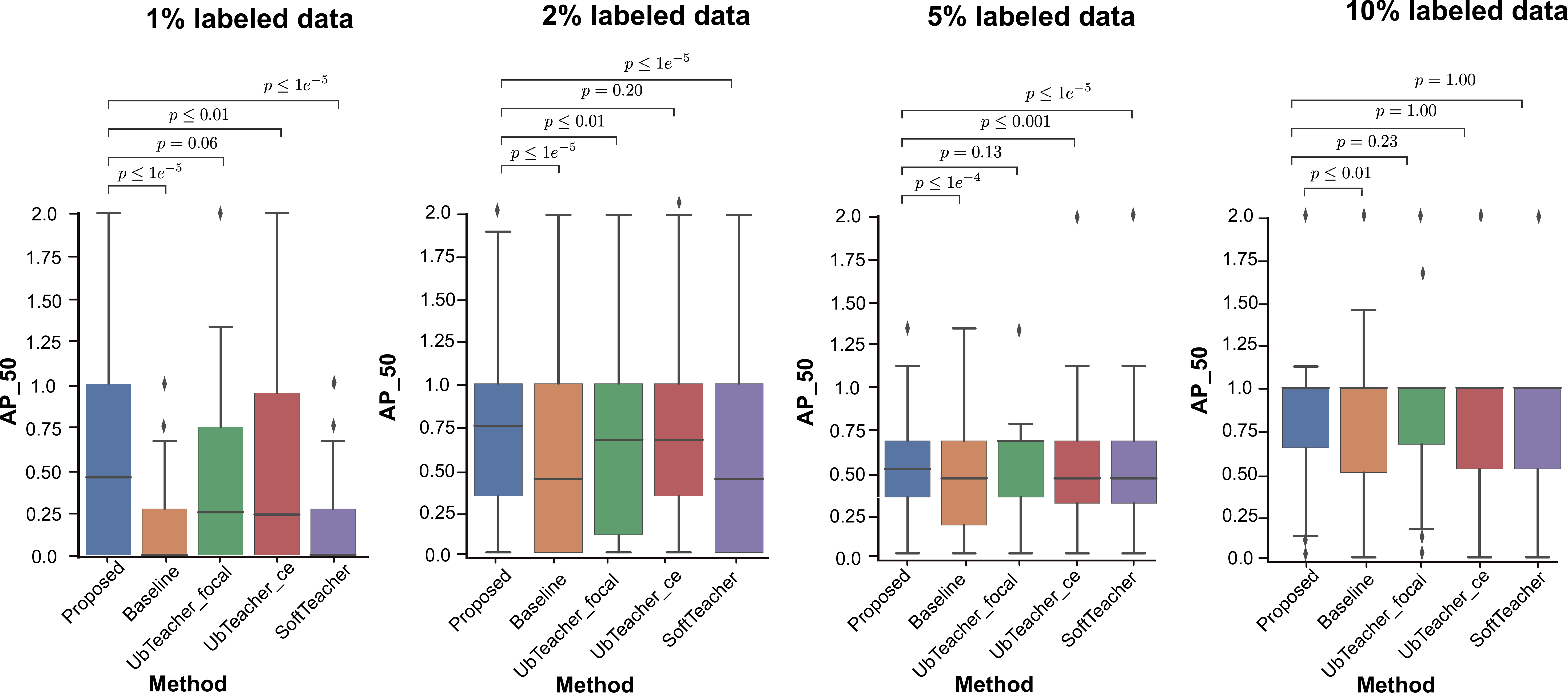}
\caption{Box-plots for paired t-test on 1\%, 2\%, 5\% and 10\% labeled data setting. Here the AP\_50 scores above 1 are only shown to represent standard deviation in the scores.} \label{fig:box-plot}
\end{figure}
\subsection{Ablation Study}\label{ablation study}
Several ablation studies were conducted to validate the effectiveness of different parameters. We evaluated the effect of initialization, confidence threshold (\(\tau\)), EMA rates and normalization parameter (s) on model performance. We trained the model with and without initialization stage and concluded that such process does improve the overall performance by a substantial margin (\textbf{Supplementary material section 7.1}). We also evaluated the model on different values of \(\tau\) where \(\tau\)=0.7 gives the best performance (\textbf{Supplementary material section 7.2}). We also performed multiple experiments to evaluate the impact of Teacher update with EMA rate on model performance for which EMA of 0.9996 gave the optimum performance (\textbf{Supplementary material section 7.3}). 

Here, we present ablation for use of different loss functions and our proposed loss with different normalization parameter `s' values. It can be observed that our proposed loss with $s=5$ provided the best performance with the highest mAP over all IoU thresholds (see Table~\ref{ab:Grid Search}).  



%
%
\begin{table}[t!]
\tbl{Normalization parameter $s$ grid search}
{\begin{tabular}{lccccc} \toprule

Loss     & \(mAP_{50:95}\)     & \(mAP_{50}\)     &  \(mAP_{75}\)     &    \(mAP_{m}\)       &     \(mAP_{l}\) \\ \midrule
Focal & 14.145 & 34.374 & 7.855 & 10.687 & 15.880 \\
\midrule
Cross Entropy &  18.008   & 42.382   & 11.387 & 13.041 & 20.135 \\
\midrule
Proposed loss (s=3)   &  16.260      &    41.438     &    8.260     &   10.801   &   18.848 \\
\midrule
Proposed loss (s=4)   &    18.475    &     44.534     &    9.252     &      13.246  &   20.483 \\
\midrule
Proposed loss (s=5)     &   \textbf{20.094}     &   \textbf{50.632}     &      \textbf{12.713}     &     \textbf{15.216}    &   \textbf{21.774}  \\
\midrule
Proposed loss (s=6)     &   18.993     &   47.597     &   11.156    &   10.380  &  22.006\\
\bottomrule
\end{tabular}}
\label{ab:Grid Search}
\end{table}

\section{Discussion and conclusion} \label{conclusion}
We demonstrate that our proposed approach performs favourably against the SOTA semi-supervised models \citep{unbiased} and \citep{softteacher} . In 1\% setting our proposed model outperforms unbiased Teacher with focal loss by a large margin and cross entropy loss by a 8 points on every evaluation metric while also outperforming SoftTeacher \citep{softteacher} model. It is worth noting that our approach achieves 50.632\% \(mAP_{50}\) on 1\% labeled data which is even higher than supervised baseline trained on 2\% labeled data and this trend can be witnessed in all settings. This improvement can be attributed to several crucial factors such as gradual improvement in pseudo label quality through EMA training which is in contrast to previous approaches in which Teacher model is freezed after training on labeled data. Another factor is the introduction of loss function which effectively increases the foreground-background distance and helps in improving detection performance.    

Furthermore, the proposed framework performs much better on \(mAP_{75}\) in all settings consistently which indicates improved localisation performance. On the 2\% labeled data setting, our model obtained 72.341\% \textit{mAP}on 50\% IoU thresholds, while unbiased Teacher on focal and cross entropy losses achieved 71.608\% and 72.416\% which is just slightly greater than our model. However, if we compare the performance of our model on \textit{mAP} at 50:95 and 75 IoU thresholds, we observe that our model consistently gives superior performance. Moreover, on 5\% and 10\% setting, Unbiased Teacher \citep{unbiased} with focal loss achieves slightly higher performance on \(mAP_{50}\), but the proposed method gives superior performance in terms of \(mAP_{50:95}\) and \(mAP_{75}\). This validates the effectiveness of our method on both classification and localization performance. We also present per class average precision (\(AP_{50:95}\)) in Table~\ref{perclassAP}. Here we observe significant improvement in average precision on all instances specially hard-to-detect classes like Specimen Bag (20 points), Irrigator (6 points) and Bipolar (16 points) against the supervised baseline.

The qualitative results also indicate strong performance of our approach as most of the tools (even when four tools in one frame) are detected and localised correctly. The localisation accuracy increases as we add more labeled data as is evident from Fig.~\ref{qualitative} from bottom to top, however the detection performance remains largely unchanged. There are some missed detections on 1\% of the labeled data setting (see row 5 in column 3) and incorrect class label prediction (see row 5 in column 4). The missed detection occurred mostly on 1\% labeled data setting where model did not see enough annotated examples. Incorrect class prediction in the bottom right may be due to less discrminative features between both instances. Similarly, the missed detection in second last image on the bottom row can be because the tool was only partly visible. \\ The paired t-test \textit{p}-values computed between the proposed method and SOTA methods are given in the Table \ref{Experimental results}. Also, we have shown a bar-plot with median and deviations and significance between the SOTA and proposed methods (see Fig.~\ref{fig:box-plot}). We can observe that our proposed approach performs well on different data settings. From Fig.~\ref{fig:box-plot}, it can be observed that for 1\% setting our method is significantly different compared to other SOTA methods with the highest median AP$_{50}$ value reported.  Similarly, on the 2\% setting our model and Unbiased Teacher model on cross entropy loss (UbTeacher\_ce) performed equally well (\textit{p}-value = 0.20) but still with the highest median value compared to other methods. Similar performance changes can be observed for 5\% data where Unbiased Teacher model on focal loss (UbTeacher\_focal) has \textit{p}-value = 0.13 (computed at AP$_{50}$) while on mAP$_{75}$ our method is still better. 
The reason behind competitive scores in these cases is because the reported APs are only done at 50\% IoU threshold, while it is evident from Table \ref{Experimental results} that our method performance for other mAPs at higher IoU thresholds has distinguishable improvements. 
However, with the 10\% labeled data setting, we reach non-significant difference in \textit{p}-values for other unbiased models and SoftTeacher model. This is because 10\% in this case is enough data for supervision during training. 

%

In this paper, we addressed a lack of annotated data problem in surgical domain for the first time by proposing a knowledge distillation framework. We tackle a multi-label, multi-class detection problem by implementing an end to end Teacher-Student learning with a multi-class foreground-background distance loss. We used strong and weak augmentation strategies to improve model robustness and class-wise NMS and EMA to improve pseudo label quality. Our experiments on m2cai16-tool dataset show the effectiveness of our model in terms of mAP on various supervision protocols against SOTA semi-supervised models. We also conducted extensive ablation experiments to demonstrate the validity of our proposed framework. \\

\vspace{-0.3cm}
\section*{Acknowledgments}
\vspace{-0.1cm}
The authors wish to thank the AI Hub and the CIIOT at ITESM for their support for carrying the experiments reported in this paper in their NVIDIA's DGX computer.


\bibliographystyle{tfcse}
\bibliography{interactcsesample}

\begin{thebibliography}{32}
\providecommand{\natexlab}[1]{#1}
\providecommand{\url}[1]{\normalfont{#1}}
\providecommand{\urlprefix}{Available from: }

\bibitem[dat(2016)]{dataset}
 2016. Tool presence detection challenge results.
  \url{http://cammau-strasbgfr/m2cai2016/indexphp/tool-presence-detection-challenge-results/}.

\bibitem[Bhatia et~al.(2007)]{bhatia2007}
Bhatia~B, Oates~T, Xiao~Y, Hu~P. 2007. Real-time identification of operating
  room state from video. In: AAAI; vol.~2. p. 1761--1766.

\bibitem[Bouget et~al.(2017)]{bouget2017}
Bouget~D, Allan~M, Stoyanov~D, Jannin~P. 2017. Vision-based and marker-less
  surgical tool detection and tracking: a review of the literature. Medical
  image analysis. 35:633--654.

\bibitem[DeVries and Taylor(2017)]{devries2017}
DeVries~T, Taylor~GW. 2017. Improved regularization of convolutional neural
  networks with cutout. arXiv preprint arXiv:170804552.

\bibitem[Hu et~al.(2017)]{agnet}
Hu~X, Yu~L, Chen~H, Qin~J, Heng~PA. 2017. Agnet: attention-guided network for
  surgical tool presence detection. In: Deep learning in medical image analysis
  and multimodal learning for clinical decision support. Springer; p. 186--194.

\bibitem[Jiang et~al.(2021)]{jiang2021semi}
Jiang~W, Xia~T, Wang~Z, Jia~F. 2021. Semi-supervised surgical tool detection
  based on highly confident pseudo labeling and strong augmentation driven
  consistency. In: Deep generative models, and data augmentation, labelling,
  and imperfections. Springer; p. 154--162.

\bibitem[Jin et~al.(2018)]{jin2018}
Jin~A, Yeung~S, Jopling~J, Krause~J, Azagury~D, Milstein~A, Fei-Fei~L. 2018.
  Tool detection and operative skill assessment in surgical videos using
  region-based convolutional neural networks. In: 2018 IEEE Winter Conference
  on Applications of Computer Vision (WACV).

\bibitem[Kranzfelder et~al.(2013)]{kranzfelder2013}
Kranzfelder~M, Schneider~A, Fiolka~A, Schwan~E, Gillen~S, Wilhelm~D,
  Schirren~R, Reiser~S, Jensen~B, Feussner~H. 2013. Real-time instrument
  detection in minimally invasive surgery using radiofrequency identification
  technology. journal of surgical research. 185(2):704--710.

\bibitem[Kurmann et~al.(2017)]{kurmann2017}
Kurmann~T, Marquez~Neila~P, Du~X, Fua~P, Stoyanov~D, Wolf~S, Sznitman~R. 2017.
  Simultaneous recognition and pose estimation of instruments in minimally
  invasive surgery. In: International conference on medical image computing and
  computer-assisted intervention. Springer. p. 505--513.

\bibitem[Lalys et~al.(2011)]{lalys2011}
Lalys~F, Riffaud~L, Bouget~D, Jannin~P. 2011. A framework for the recognition
  of high-level surgical tasks from video images for cataract surgeries. IEEE
  Transactions on Biomedical Engineering. 59(4):966--976.

\bibitem[Liu et~al.(2021)]{unbiased}
Liu~YC, Ma~CY, He~Z, Kuo~CW, Chen~K, Zhang~P, Wu~B, Kira~Z, Vajda~P. 2021.
  Unbiased teacher for semi-supervised object detection. In: Proceedings of the
  International Conference on Learning Representations (ICLR).

\bibitem[Mishra et~al.(2017)]{mishra2017}
Mishra~K, Sathish~R, Sheet~D. 2017. Learning latent temporal connectionism of
  deep residual visual abstractions for identifying surgical tools in
  laparoscopy procedures. In: Proceedings of the IEEE Conference on Computer
  Vision and Pattern Recognition Workshops. p. 58--65.

\bibitem[Raju et~al.(2016)]{raju2016m2cai}
Raju~A, Wang~S, Huang~J. 2016. M2cai surgical tool detection challenge report.
  In: Workshop and Challenges on Modeling and Monitoring of Computer Assisted
  Intervention (M2CAI), Athens, Greece, Technical report. p. 1--4.

\bibitem[Reiter et~al.(2012)]{reiter2012}
Reiter~A, Allen~PK, Zhao~T. 2012. Articulated surgical tool detection using
  virtually-rendered templates. In: Computer Assisted Radiology and Surgery
  (CARS). p. 1--8.

\bibitem[Russakovsky et~al.(2015)]{imagenet}
Russakovsky~O, Deng~J, Su~H, Krause~J, Satheesh~S, Ma~S, Huang~Z, Karpathy~A,
  Khosla~A, Bernstein~M, et~al. 2015. Imagenet large scale visual recognition
  challenge. International journal of computer vision. 115(3):211--252.

\bibitem[Sahu et~al.(2016)]{sahu2016tool}
Sahu~M, Mukhopadhyay~A, Szengel~A, Zachow~S. 2016. Tool and phase recognition
  using contextual cnn features. arXiv preprint arXiv:161008854.

\bibitem[Sarikaya et~al.(2017)]{sarikaya2017}
Sarikaya~D, Corso~JJ, Guru~KA. 2017. Detection and localization of robotic
  tools in robot-assisted surgery videos using deep neural networks for region
  proposal and detection. IEEE transactions on medical imaging.
  36(7):1542--1549.

\bibitem[Shi et~al.(2020{\natexlab{a}})]{shi2020}
Shi~P, Zhao~Z, Hu~S, Chang~F. 2020{\natexlab{a}}. Real-time surgical tool
  detection in minimally invasive surgery based on attention-guided
  convolutional neural network. IEEE Access. 8:228853--228862.

\bibitem[Shi et~al.(2020{\natexlab{b}})]{shi2020:IEEE}
Shi~P, Zhao~Z, Hu~S, Chang~F. 2020{\natexlab{b}}. Real-time surgical tool
  detection in minimally invasive surgery based on attention-guided
  convolutional neural network. IEEE Access. 8:228853--228862.

\bibitem[Sohn et~al.(2020)]{sohn2020fixmatch}
Sohn~K, Berthelot~D, Carlini~N, Zhang~Z, Zhang~H, Raffel~CA, Cubuk~ED,
  Kurakin~A, Li~CL. 2020. Fixmatch: Simplifying semi-supervised learning with
  consistency and confidence. Advances in Neural Information Processing
  Systems. 33:596--608.

\bibitem[Speidel et~al.(2009)]{speidel2009}
Speidel~S, Benzko~J, Krappe~S, Sudra~G, Azad~P, M{\"u}ller-Stich~BP, Gutt~C,
  Dillmann~R. 2009. Automatic classification of minimally invasive instruments
  based on endoscopic image sequences. In: Medical Imaging 2009: Visualization,
  Image-Guided Procedures, and Modeling; vol. 7261. International Society for
  Optics and Photonics. p. 72610A.

\bibitem[Twinanda et~al.(2016{\natexlab{a}})]{twinanda2016}
Twinanda~AP, Mutter~D, Marescaux~J, de~Mathelin~M, Padoy~N. 2016{\natexlab{a}}.
  Single-and multi-task architectures for tool presence detection challenge at
  m2cai 2016. arXiv preprint arXiv:161008851.

\bibitem[Twinanda et~al.(2016{\natexlab{b}})]{endonet}
Twinanda~AP, Shehata~S, Mutter~D, Marescaux~J, De~Mathelin~M, Padoy~N.
  2016{\natexlab{b}}. Endonet: a deep architecture for recognition tasks on
  laparoscopic videos. IEEE transactions on medical imaging. 36(1):86--97.

\bibitem[Van~Engelen and Hoos(2020)]{van2020}
Van~Engelen~JE, Hoos~HH. 2020. A survey on semi-supervised learning. Machine
  Learning. 109(2):373--440.

\bibitem[Vardazaryan et~al.(2018)]{vardazaryan2018}
Vardazaryan~A, Mutter~D, Marescaux~J, Padoy~N. 2018. Weakly-supervised learning
  for tool localization in laparoscopic videos. In: Intravascular imaging and
  computer assisted stenting and large-scale annotation of biomedical data and
  expert label synthesis. Springer.

\bibitem[Wang et~al.(2019)]{wang2019graph}
Wang~S, Xu~Z, Yan~C, Huang~J. 2019. Graph convolutional nets for tool presence
  detection in surgical videos. In: International Cofnference on Information
  Processing in Medical Imaging. Springer. p. 467--478.

\bibitem[Ward et~al.(2021)]{ward2021computer}
Ward~TM, Mascagni~P, Ban~Y, Rosman~G, Padoy~N, Meireles~O, Hashimoto~DA. 2021.
  Computer vision in surgery. Surgery. 169(5):1253--1256.

\bibitem[Wu et~al.(2019)]{wu2019detectron2}
Wu~Y, Kirillov~A, Massa~F, Lo~WY, Girshick~R. 2019. Detectron2;
  [\url{https://github.com/facebookresearch/detectron2}].

\bibitem[Xu et~al.(2021)]{softteacher}
Xu~M, Zhang~Z, Hu~H, Wang~J, Wang~L, Wei~F, Bai~X, Liu~Z. 2021. End-to-end
  semi-supervised object detection with soft teacher. In: Proceedings of the
  IEEE/CVF International Conference on Computer Vision. p. 3060--3069.

\bibitem[Yang et~al.(2021)]{yang2021efficient}
Yang~Y, Zhao~Z, Shi~P, Hu~S. 2021. An efficient one-stage detector for
  real-time surgical tools detection in robot-assisted surgery. In: Annual
  Conference on Medical Image Understanding and Analysis. Springer. p. 18--29.

\bibitem[Yoon et~al.(2020)]{yoon2020semi}
Yoon~J, Lee~J, Park~S, Hyung~WJ, Choi~MK. 2020. Semi-supervised learning for
  instrument detection with a class imbalanced dataset. In: Interpretable and
  annotation-efficient learning for medical image computing. Springer; p.
  266--276.

\bibitem[Zhang et~al.(2020)]{zhang2020}
Zhang~B, Wang~S, Dong~L, Chen~P. 2020. Surgical tools detection based on
  modulated anchoring network in laparoscopic videos. IEEE Access.
  8:23748--23758.

\end{thebibliography}

\newpage {\Large \section{\Large Supplementary material}}

\subsection{Effect of initialisation}

It is essential to have a good initialisation to the student-teacher mutual learning in order to better generate optimal pseudo-boxes. We provide a comparison between learning with and without initialisation as shown in Fig.. A good initialisation helps the teacher to generate good quality pseudo labels at the start, which then boost initial performance and the model converges at a higher accuracy as compared to the model without initialisation.

\subsection{Effect of the confidence threshold}

We apply confidence thresholding to filter out low-confidence bounding boxes, which are most likely to be false-positives. The number of generated pseudo boxes will decrease if we vary the threshold downwards from 0.9 to 0. In Fig. 2 we show the variation in validation performance at various values of the threshold \(\tau\) \(\in \) \{0.9, 0.8, 0.7, 0.6\}. At very high value of threshold (\(\tau\) = 0.9), model can not perform satisfactorily since it generates very few pseudo boxes. As can be seen in Fig. 3, the model misses out an instance which could have lower prediction confidence. At a low threshold (\(\tau\) = 0.6), the model generates too many pseudo boxes, which are would contain many false-positive instances, thus degrading the overall performance.  The best results are obtained when \(\tau\) is set to 0.7.

\begin{figure}[h!]
    \captionsetup{justification=justified}
    \subfloat[\centering]{{\includegraphics[scale=0.52]{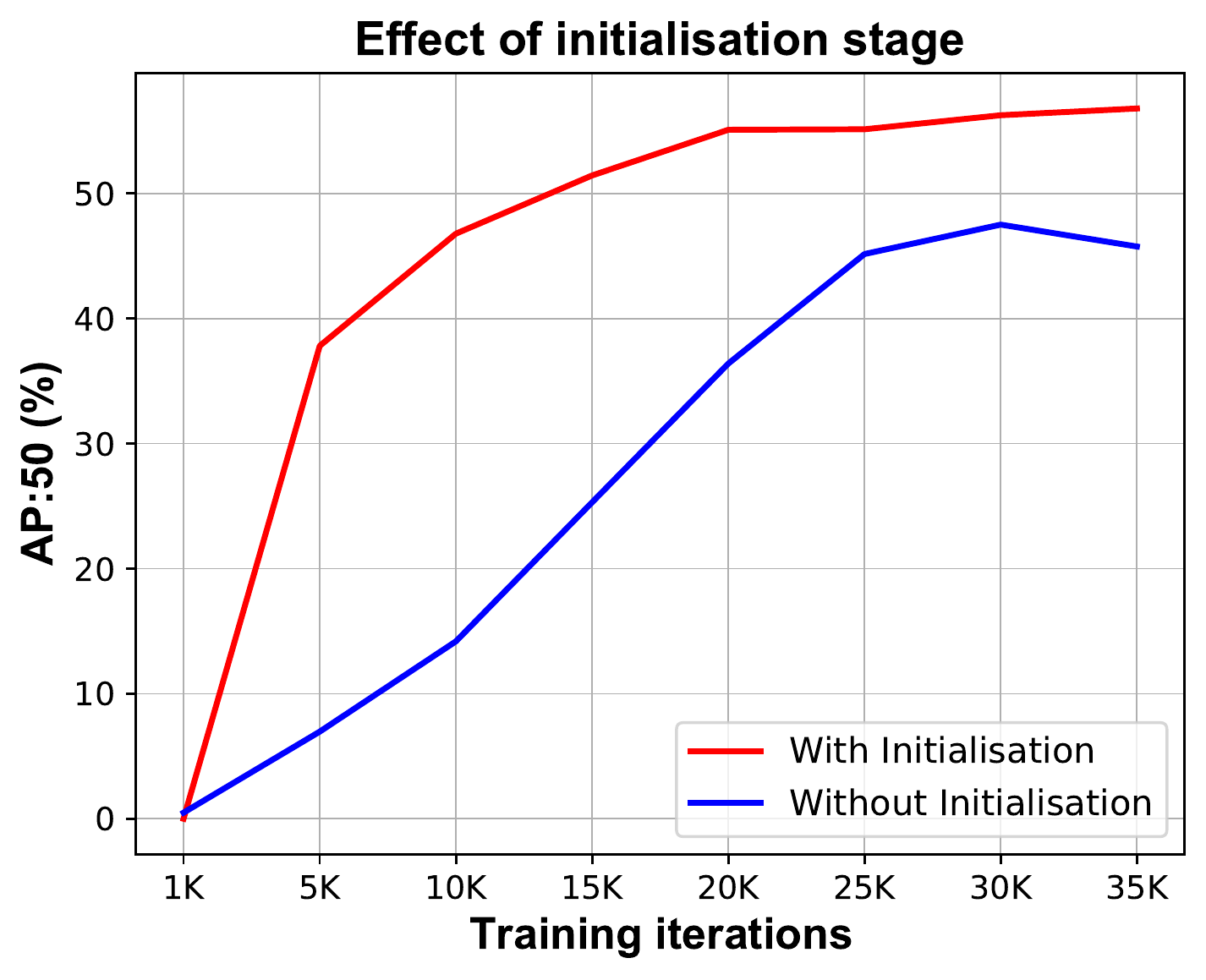} }}%
    \subfloat[\centering]{{\includegraphics[scale=0.52]{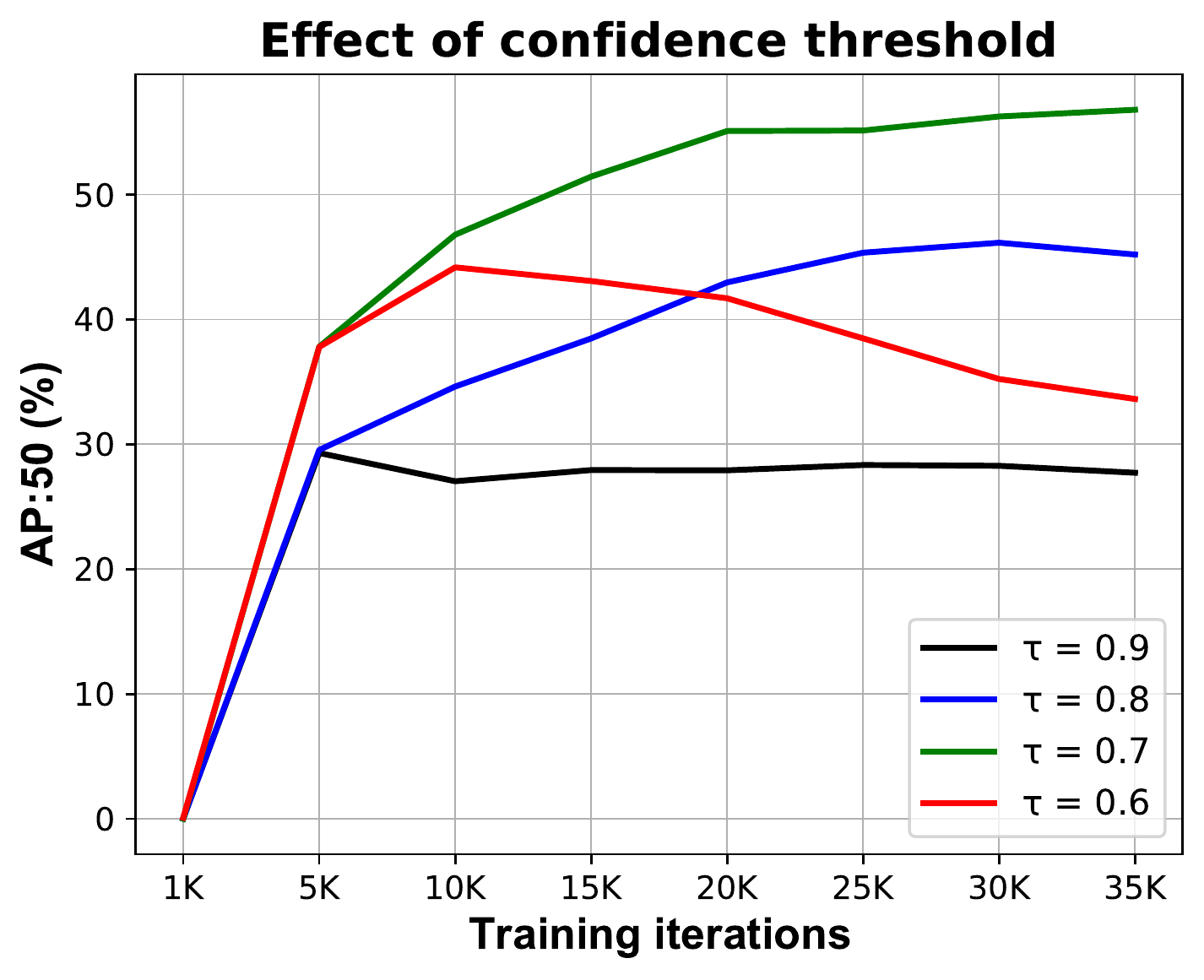}}}%
    \caption{\textbf{Effect of initialization stage and different confidence thresholds on mAP.} (a) Validation mAP w/without initialisation  (b) Validation mAP variation with \(\tau\).}%
    \label{initalization and confidence}%
\end{figure}

\begin{figure}[h!]
\captionsetup{justification=justified}
\centering
\includegraphics[width=5in ,height=4in, keepaspectratio]{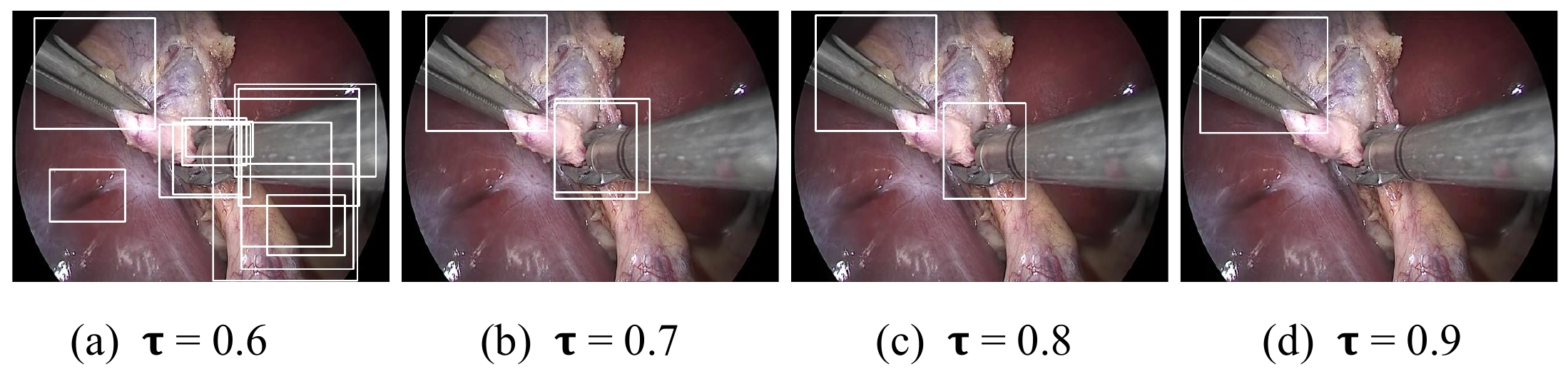}
\caption{Visualization of predictions on various values of \(\tau\). Higher values of \(\tau\) result in greater precision, but lower recall, as seen one instance is missed at \(\tau=0.9.\)  } \label{ab:confidence bounding boxes}
\end{figure}

\subsection{Effect of modifying the EMA rate}
We evaluate teacher model on different EMA rates to determine how it impacts its performance Fig. 4. At a small value (\(\alpha\)=0.5), the model exhibits a lower mAP and high variance, since student model contribution in the teacher update is higher (Eq 4). This has the effect that the teacher gets indirectly affected by the noisy pseudo label trained student. As \(\alpha\) is increased,  the teacher performance is stabilized. The model yields the best result on \(\alpha\)=0.9996, but if the \(\alpha\) is increased further (eg. 0.9999), it will become inordinately slow as the teacher model gets its weights from the weights of previous iteration teacher model.   

\begin{figure}[h!]
\centering
\captionsetup{justification=justified}

\includegraphics[width= 4in, height=2.7in, keepaspectratio]{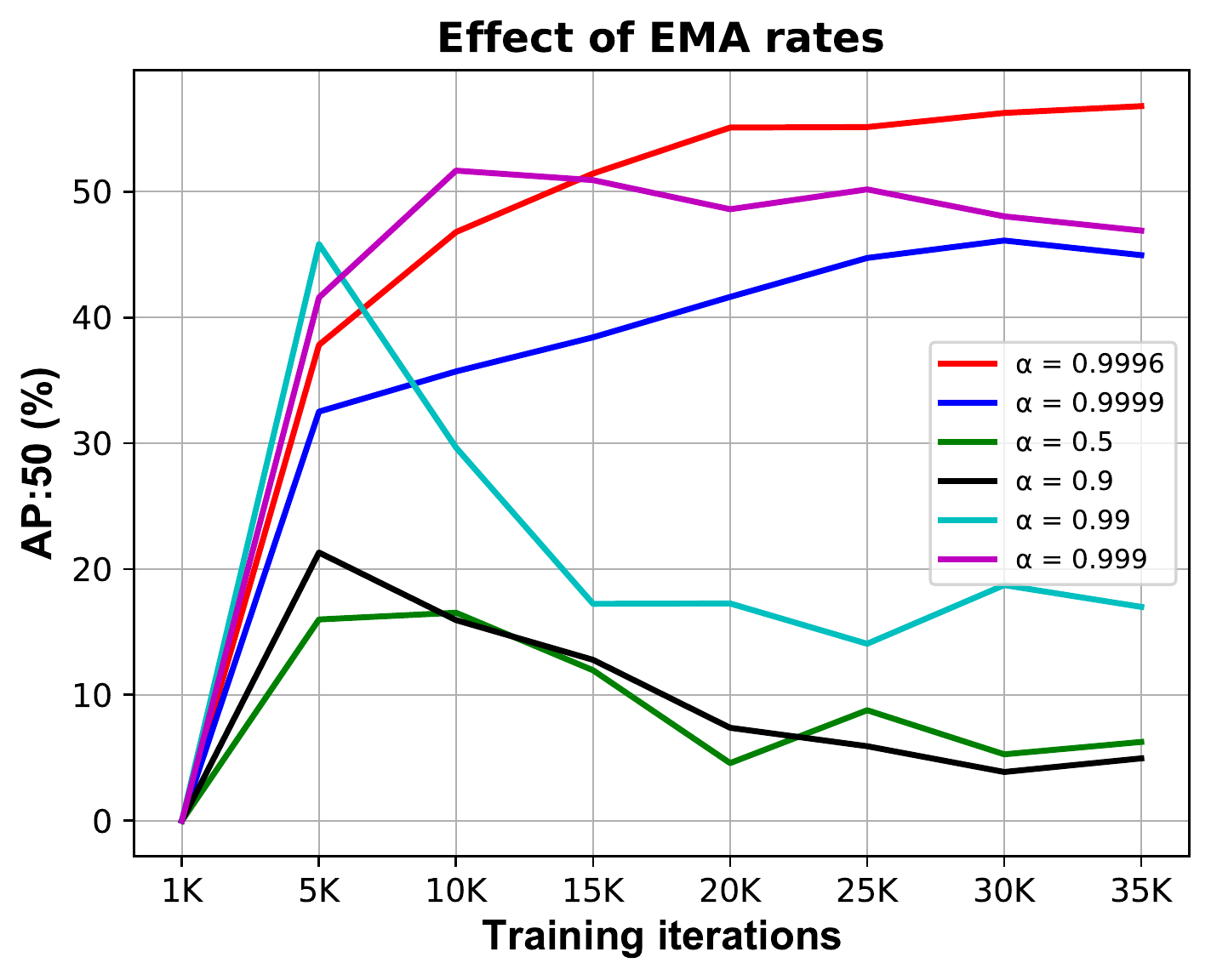}
\caption{ Validation mAP on teacher network with various EMA rates (\(\alpha\)). With small \(\alpha\) (eg. 0.5), model has very low mAP and larger variance, but as \(\alpha\) increases (eg. 0.999), mAP of model gradually improves and variance reduces. Best mAP is obtained at \(\alpha\)=0.9996.} \label{ab:effect of ema}
\end{figure}

\end{document}